\documentclass[letterpaper, 10 pt, conference]{ieeeconf}  % Comment this line out if you need a4paper
\usepackage{graphicx}
\usepackage{multirow}

\usepackage{amsmath}
\usepackage{amssymb}

\IEEEoverridecommandlockouts                              % This command is only needed if 
\title{\LARGE \bf
A Unified Multi-Task Learning Framework of Real-Time Drone Supervision for Crowd Counting
}

\author{
Siqi Gu and Zhichao Lian*\\% <-this % stops a space
School of Computer Science and Engineering\\Nanjing University of Science and Technology\\Nanjing, China

\thanks{*This work was not supported by any organization}% <-this % stops a space
}

\begin{document}

\maketitle
\thispagestyle{empty}
\pagestyle{empty}

%%%%%%%%%%%%%%%%%%%%%%%%%%%%%%%%%%%%%%%%%%%%%%%%%%%%%%%%%%%%%%%%%%%%%%%%%%%%%%%%
\begin{abstract}

In this paper, a novel Unified Multi-Task Learning Framework of Real-Time Drone Supervision for Crowd Counting (MFCC) is proposed, which utilizes an image fusion network architecture to fuse images from the visible and thermal infrared image, and a crowd counting network architecture to estimate the density map. The purpose of our framework is to fuse two modalities, including visible and thermal infrared images captured by drones in real-time, that exploit the complementary information to accurately count the dense population and then automatically guide the flight of the drone to supervise the dense crowd. To this end, we propose the unified multi-task learning framework for crowd counting for the first time and re-design the unified training loss functions to align the image fusion network and crowd counting network. We also design the Assisted Learning Module (ALM) to fuse the density map feature to the image fusion encoder process for learning the counting features. To improve the accuracy, we propose the Extensive Context Extraction Module (ECEM) that is based on a dense connection architecture to encode multi-receptive-fields contextual information and apply the Multi-domain Attention Block (MAB) for concerning the head region in the drone view. Finally, we apply the prediction map to automatically guide the drones to supervise the dense crowd. The experimental results on DroneRGBT data set show that, compared with the existing methods, ours has comparable results on objective evaluations and an easier training process.

\end{abstract}

%%%%%%%%%%%%%%%%%%%%%%%%%%%%%%%%%%%%%%%%%%%%%%%%%%%%%%%%%%%%%%%%%%%%%%%%%%%%%%%%
\section{INTRODUCTION}

Recently, drones have emerged in a wide range of applications, such as visual surveillance, rescue and entertainment. Although fixed cameras are widely used to monitor and detect crowd density in crowded indoor places, such as shopping malls and subway stations, the images captured by fixed cameras often have the phenomenon of large human crowds cover each other or disruptive backgrounds. In outdoor places, crowd analysis methods used from drone view could be more convenient, accurate and flexible to grasp the crowd movement and crowd density situation. Thus, automated crowd analysis from drones has been an increasingly hot topic in computer vision and attracted much attention.

To begin with, due to strong occlusions, distortions of perspective, scale variations and different human mass, identification and counting of people in the dense crowd are more challenging. Diverse CNN-based approaches have been proposed to address the crowd counting in the complex scene. The classic scale-aware CNN-based models handle the scale variation problem by taking advantage of multi-column or multi-resolution architectures, like \cite{mcnn}. Another set of approaches concerned local and global contextual information which is extracted from CNN-based frameworks to receive lower estimation errors including \cite{VLAD}, \cite{scar} and \cite{csr}. Recently, Wang et al.\cite{DM} used Optimal Transport (OT) to measure the similarity between the normalized predicted density map and the normalized ground truth density map to decrease generalization error. Wan and Chen\cite{noisy} concerned more about the annotation noise in crowd counting to get adaptive results on annotation noise. Zhou et al. \cite{local-aware} proposed a simple but effective locality-based learning paradigm to produce generalizable features by alleviating sample bias.

Furthermore, images and videos captured by drones have different camera perspectives and low resolution compared with traditional data sets which would make the task of crowd counting and detection more difficult. Researchers have proposed several improved methods to solve the scale changes during the feature extracted and density map prediction process \cite{drone-scnet}\cite{geometric}\cite{scale-adaptive}\cite{soft-csrnet}. In addition, other researchers \cite{drone1}\cite{drone2}\cite{drone3} have put forward frameworks to not only detect and count crowds but also to easily track and monitor big crowds by using drones.

However, conventional crowd counting data sets contain only visible images, which always has drawbacks of illumination changes and poor imaging conditions in the nighttime or shadows. But the thermal infrared data can perfectly complement that and allow the effectiveness of crowd counting methods both day and night. Recently, Peng et al. has proposed a drone-based RGB-Thermal crowd counting data set DroneRGBT\cite{RGBT}, which also proposed a multi-modal crowd counting network (MMCCN) to utilize the multi-modal inputs for crowd counting. But it required the complex aligner model and huge computation cost because it had to train multiple branches which increased the difficulty of training. 

Intuitively, we suppose that building the framework which connect the image fusion network and crowd counting network in stages will obtain great results. Nevertheless, we find that it is difficult to achieve prediction map with high quality because the original loss functions used in fusion network do not contain the counting features.Therefore, in this work we propose the unified learning framework to estimate the density map, which can easily reuse the existing fusion model like \cite{RFN}\cite{nestfuse}\cite{Did} and align it with the crowd counting network using unified loss functions. Thus we add the Assisted Learning Module (ALM) to fuse the density map information during the fusion encoding process and re-design the decomposed loss function to develop the counting residual fusion network architecture (CRFN-Nest) based on RFN-Nest\cite{RFN}. 

\begin{figure}[t]
      \centering
      \framebox{\parbox[c][4cm][c]{3.2in}{
      \includegraphics[scale=0.28]{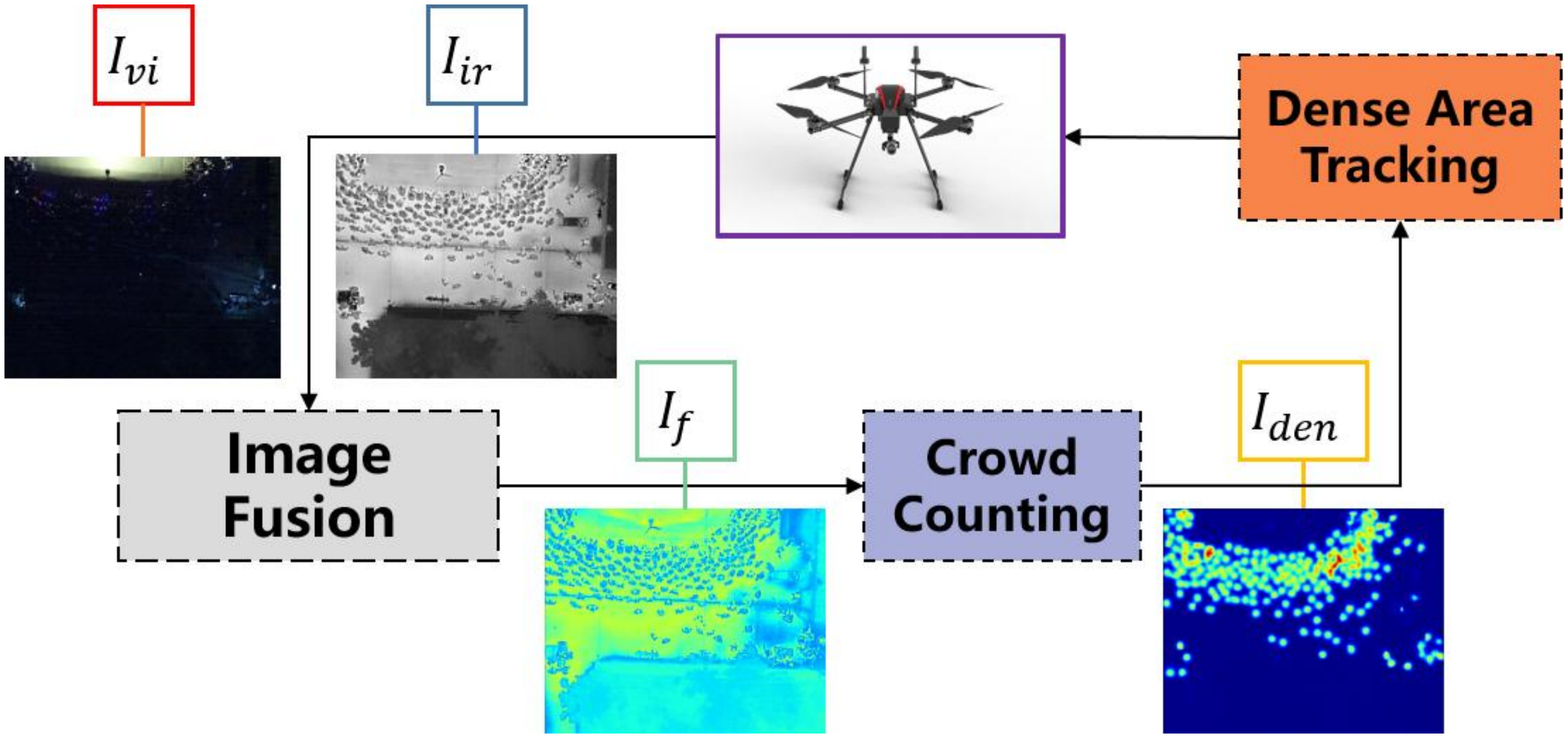}}}
      \caption{The flow chart of the end-to-end Multi-Task Fusion Counting Framework (MFCC)}
      \label{xitongtu}
   \end{figure}

What’s more, to increase the accuracy, we also propose the multi-receptive-fields crowd counting machine to extract multi-scale context with robust high-level feature representation from fusion images. Specifically, we propose the Extensive Context Extraction Module (ECEM) to build connections between the dilated layers with different dilated rates. Also, to screen the crowd counting features we introduce the Multi-domain Attention Block (MAB) to encode the spatial and channel dependencies in the whole feature map, which uses long-range dependency to reduce the error estimation for background and redundancy under the counting scene. Finally, we apply the prediction map generated by MFCC to automatically guide drones to supervise high dense crowds via designing a real-time dense area supervising method. The flow chart of our proposal is shown in Fig.\ref{xitongtu}.

% ------------------------------------------------
\section{Proposed Method}
\begin{figure*}[t]
      \centering
      \framebox{\parbox[c][7.0cm][c]{5.5in}{
      \includegraphics[scale=0.45]{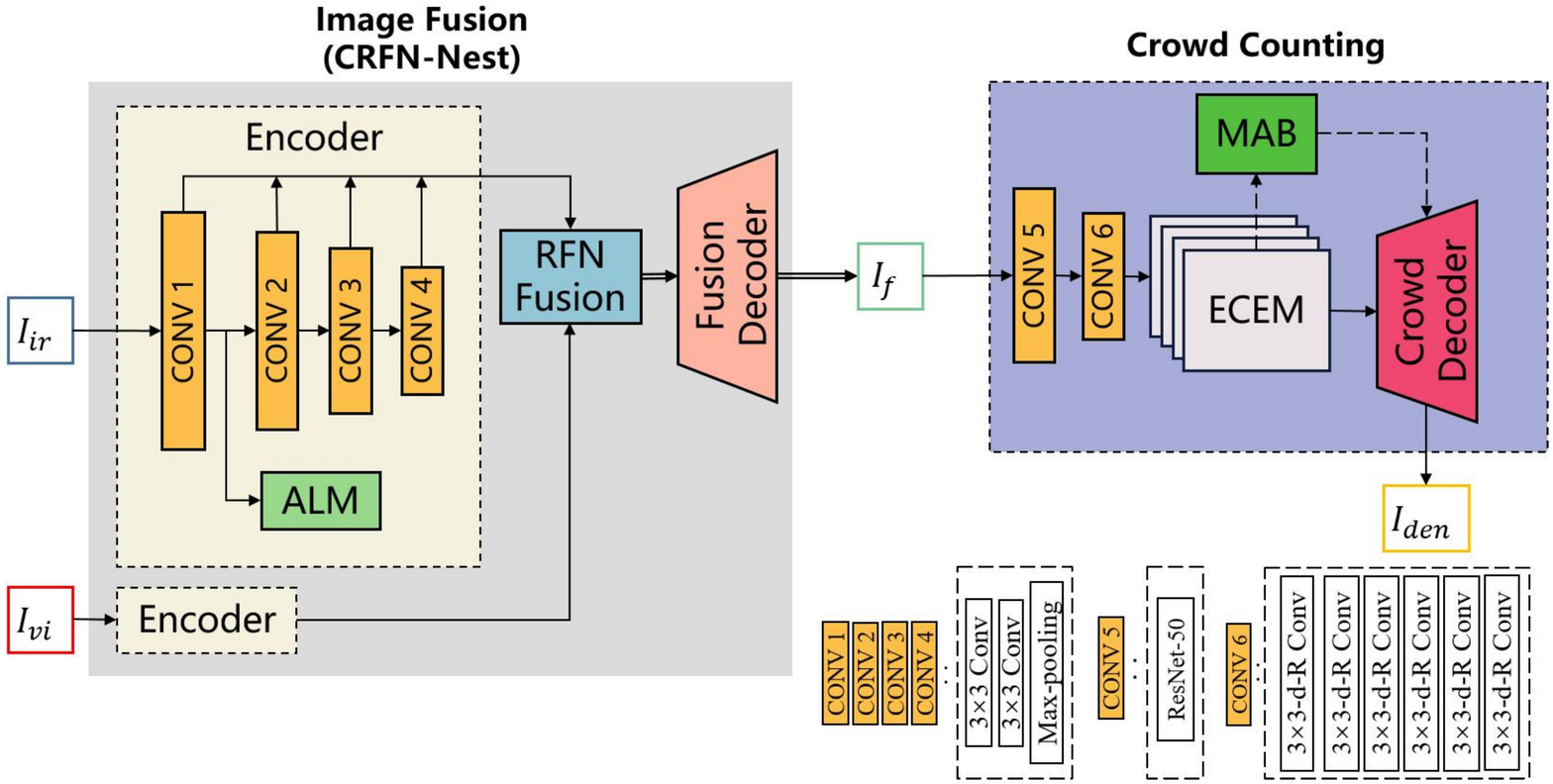}}}
      \caption{The pipeline of Multi-Task Fusion Counting Network. '3$\times$3' represents the normal convolutional operation. '3$\times$3-d-R' represents the convolutional operation with kernel size of 3$\times$3, dilation rate of 2. The 'R' means that the ReLU layer is added to this convolutional layer}
      \label{wholemodel}
   \end{figure*}

\begin{figure}[t]
      \centering
      \framebox{\parbox[s][4.2cm][c]{2.33in}{\includegraphics[scale=0.20]{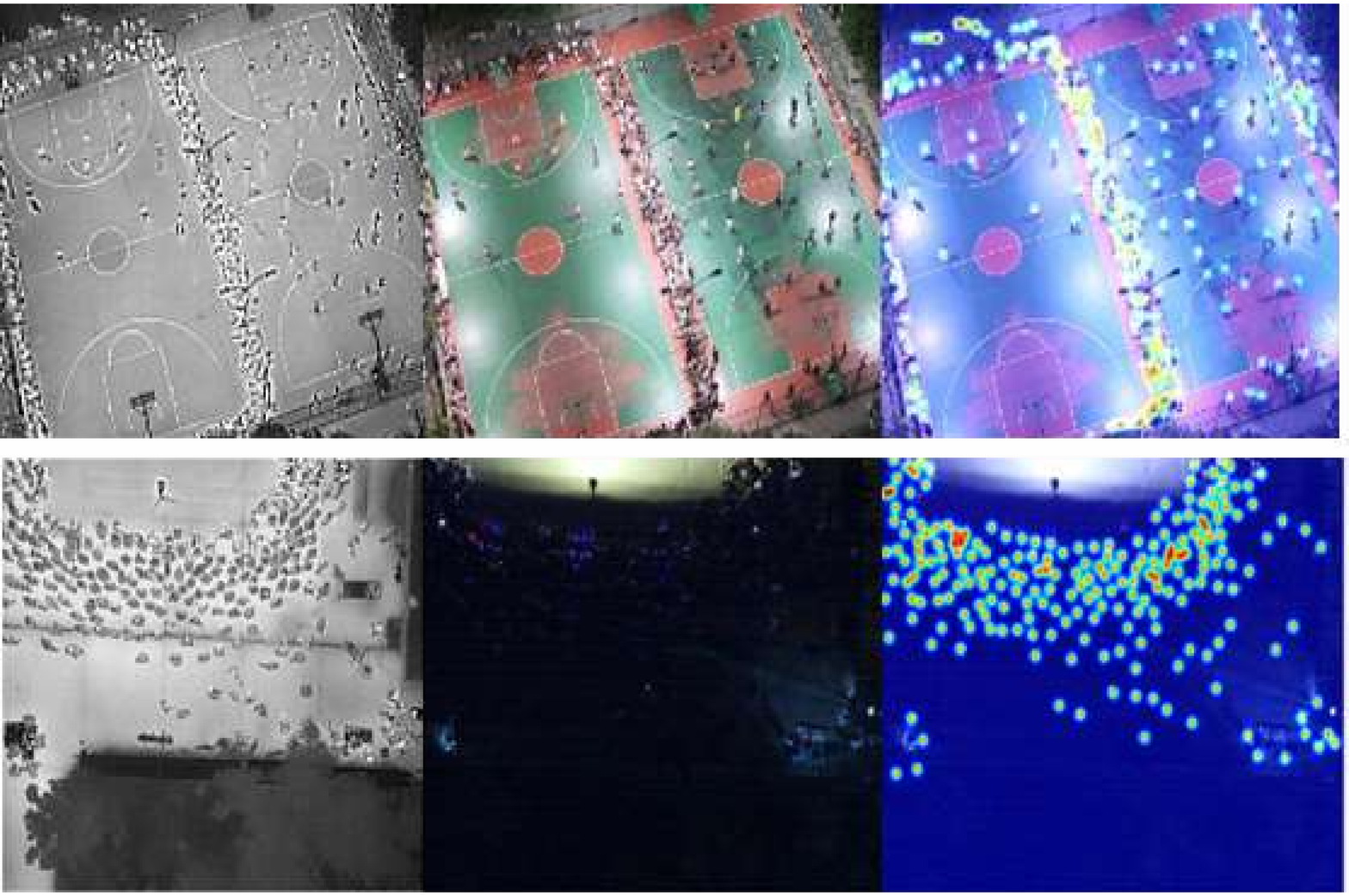}}}
      \caption{Image pairs and ground truth from DrondRGBT data set}
      \label{duomotai}
   \end{figure}
   
This chapter introduces the details of the proposed unified multi-task learning framework for Crowd Counting. Different from the previous crowd counting methods, we fuse the visible images and thermal infrared images on the feature-level to make it keep the complementary information from the multi-modal inputs. Then we send the fusion image into the crowd counting network to estimate the prediction map. Also, for building the end-to-end unified learning framework we re-design the loss function during the training process. We propose the MFCC not only utilize the multi-modal inputs to increase the accuracy of the counting prediction map but also propose a flexible framework to apply different image fusion and crowd counting strategies. The pipeline of the Multi-Task Fusion Counting Network is shown in Fig.\ref{wholemodel}. In this section, we describe the flow chart of whole network and then present the details of each submodules: ALM, ECEM, and MAB.

\subsection{The Architecture of MFCC}
\subsubsection{CRFN-Nest}
As shown in Fig.\ref{duomotai}, regardless of light conditions, we prefer the detailed information from thermal infrared images and background information from visible images to better estimate the crowd counting. Inspired by RFN-Nest\cite{RFN}, we improve the image fusion model by adding the Assisted Learning Module (ALM) to better reconstruction counting features. The pair of images are separately fed in two encoder modules to extract multi-scale features and then send to the same RFN-Fusion module to fuse the multi-modal features, and finally be reconstructed by the same reconstruction decoder module. As shown in Fig.\ref{wholemodel}, $I_{ir}$ and $I_{vi}$ indicate the source images (infrared image and visible image). $I_f$ denotes the output of CRFN-Nest, which is the fused image. The architecture of the encoder in CRFN-Nest is constituted by four RFN networks in the RFN Fusion module and these RFN networks share the same architecture but with different weights, we set it following RFN-Nest. The framework of CRFN-Nest is shown in the grey box in Fig.\ref{wholemodel}.

However, the loss function set in RFN-Nest could not orientate the fusion image towards the crowd counting task. Therefore, we design the Assisted Learning Module (ALM) to further align the density map features to the encoder backward process. Considering the fusion task and counting task are both based on the regression model, we transform the regression loss to classification one by applying the assisted learning method because it brings sharper linear loss and supplementary characteristic information to force the fusion model to make trade-offs in the direction required by the counting task. This module will not be used during the validation process.

\subsubsection{ECEM}
\begin{figure}[t]
      \centering
      \framebox{\parbox[c][2.8cm][c]{3.2in}{\includegraphics[scale=0.33]{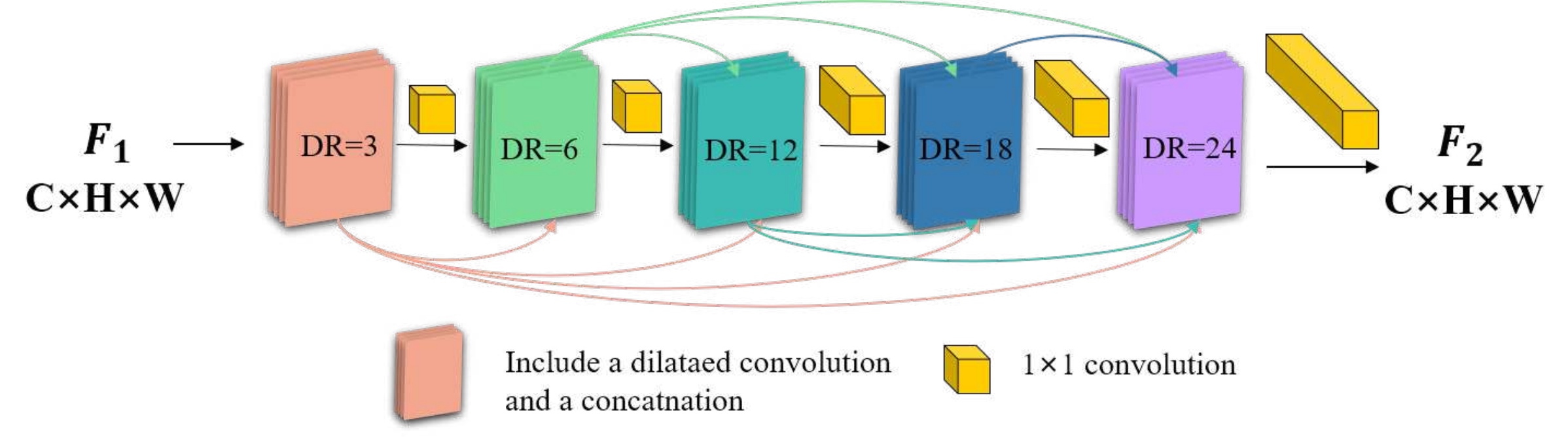}}}
      \caption{The detailed architectures of the Extensive Context Extraction Module (ECEM) in EBM-Net. 'DR' represents the dilated rate of the current dilated convolution.}
      \label{ecem}
   \end{figure}

The previous scale-aware methods are not robust enough to deal with large variations in scales and spatial at the same time, especially from drones. Furthermore, non-uniform density in a single image or data set is prone to mislead the existing models because of the similar class-specific responses. Thus we design the Extensive Context Extraction Module (ECEM) to extract the feature from the fusion image.
We use ResNet \cite{Res50} and dilatation module as the front-end of our method inspired by CSRNet \cite{csr} and Spatial-/Channel-wise Attention Regression Networks for Crowd Counting (SCAR) \cite{scar}. It enlarges the respective field of the extracted feature map and outputs a 64-D channel with 1/8 size feature maps. Different from the method proposed in \cite{Attention-guided}, we remove a global max-pooling layer, a 1×1 convolution, and a bilinear interpolation on the sampling layer to better build the background modeling in the counting task and increase the generalization ability to fit the large-range scale changes. 
We change the hyper-parameters to keep the size of the channel unchanged. To be specific, as shown in Fig.\ref{ecem}, after obtaining the feature map (i.e., $F_1$) from preceding layers we feed it into our ECEM. The size of the backbone’s output is $C\times H\times W$. We utilize five dilated convolutional layers with different rates, e.g., rate = 3, 6, 12, 18, 24. In addition, dense connections are employed to our ECEM, where the results of each dilated layer are integrated with the former ones and then are feed to the later ones. DenseNet \cite{DenseNet} designs the dense connection to handle the problems of vanishing gradients and strengthens features propagation when the CNN model is quite deep. Concatenation operation is applied to fuse features and integrate the information. As a result, these separated convolutional layers can harvest multiple feature maps in various-scale and large-range receptive fields. A semantically informative and globally distributed feature map (i.e., $F_2$) is generated after our ECEM and the size of it is the same as input, which means we keep the global information without deepening the channel of the feature map. 

\subsubsection{MAB}
Since the feature map generated by ECEM is full of rich receptive field information and contextual information, not all of them are useful to facilitate the performance of crowd counting. Inspired by Non-local block\cite{Non-local}, we want to weigh the sum of the original local features map to learn this density change in spatial. Specifically, the class-specific response includes two types: foreground (head region) from infrared features and background (other regions) from visible features. The large-range dependencies on spatial dimension features tend to mislead the model because the foreground’s textures are very similar to those of some background regions (trees, roads, and so on). Therefore, we aim to embed the spatial-wise and channel-wise attention models to reduce these problems which we named Multi-domain Attention Block (MAB).

\begin{figure}[t]
      \centering
      \framebox{\parbox[c][6cm][c]{3.1in}{\includegraphics[scale=0.32]{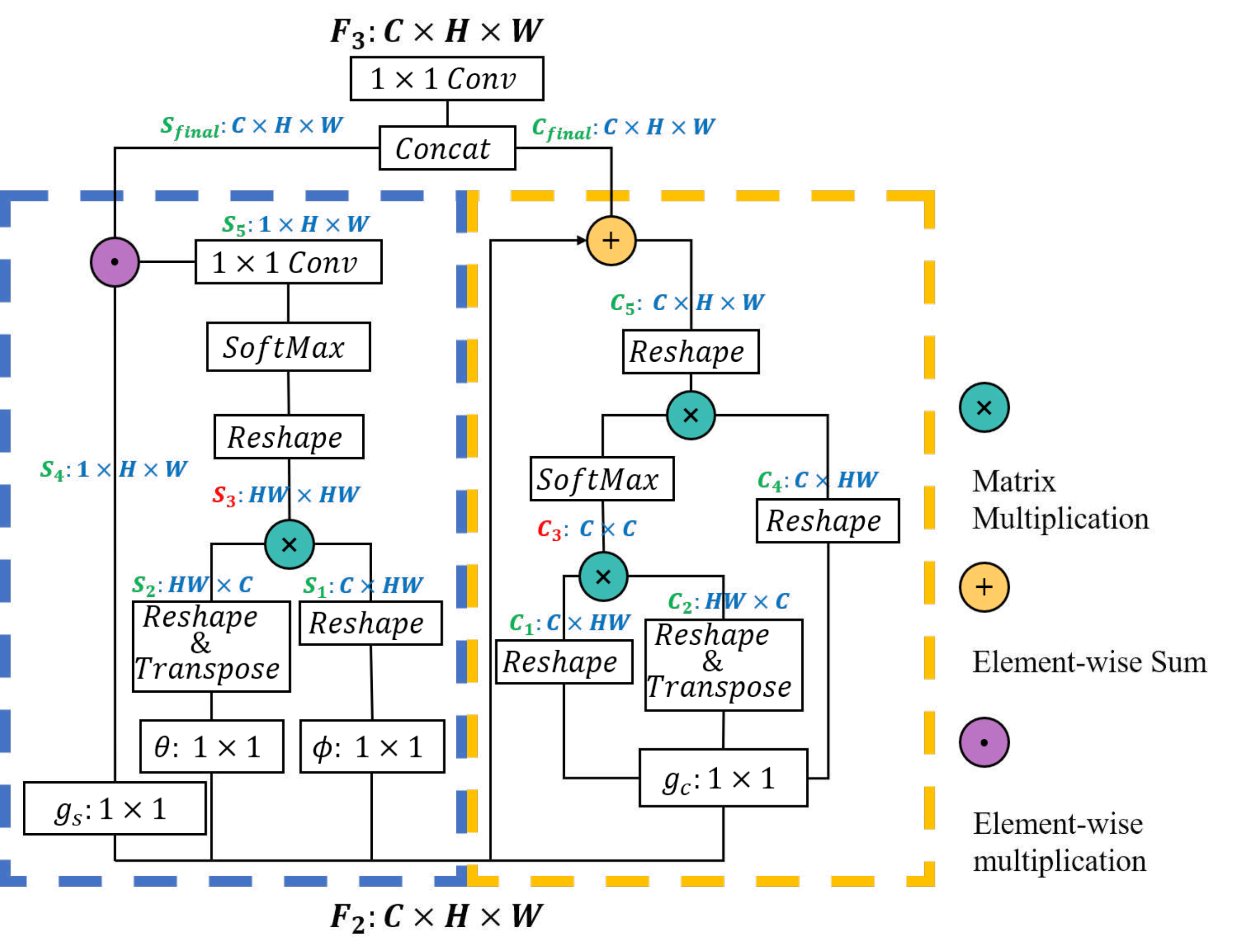}}}
      \caption{ The detailed architectures of the Multi-domain Attention Block (MAB).}
      \label{mab}
   \end{figure}

The detailed architectures of the MAB are shown in Fig.\ref{mab}. The spatial-wise attention model is described in the blue box and the channel-wise attention model is described in the yellow box. The output of ECEM with the size of $C\times H\times W$ is fed into several $1\times1$ convolutional layers. Then by the reshape or transpose operations, the spatial or the channel information is kept respectively. 

In the spatial domain, we adaptive several operations to pay more attention to the relations between subregions which are more relevant. Consequently, the feature map keeps more clear semantics and contextual dependencies within regions. 
In the channel domain, we regard each channel map of the high-level features as a class-specific response, and different semantic responses are related to each other. The channel-wise attention module significantly prompts the regression result and reduces the error prediction for backgrounds. 
Following the generic non-local operation in deep neural networks, the attention mechanism can be defined as:

%----------------------formula 1--------------------------------------------------
\begin{equation}
\begin{split}
     y_i=\ \frac{1}{C(x)}\ \sum_{\forall j}{f(x_i,\ x_j)g(x_j)}
\end{split}
\end{equation}
%------------------------------------------------------------------------

Here $i$ is the index of an output position whose response is to be computed and $j$ is the index that enumerates all possible positions. $x$ represents features and $y$ is the output signal of the same size as $x$. Function $f$ computes the affinity relationship between $i$ and all $j$. Function g computes a representation of the input signal at the position $j$. The response is normalized by a factor $C(x)$.

Several versions of $f$ and $g$ can be used. Here we consider $g$ in the form of a linear embedding and utilize two kinds of $f$:
%----------------------formula 2345--------------------------------------------------
\begin{equation}
     g\left(x_j\right)=W_gx_j \\
\end{equation}
\begin{equation}
     f(x_i, x_j)\ =\ e^{x_is^Tx_j}\\
\end{equation}
\begin{equation}
f\left(x_i,x_j\right)=e^{{\theta\left(x_i\right)}^T\phi\left(x_j\right)}
\end{equation}
\begin{equation}
\begin{split}
\theta(x_i)=W_\theta x_i, \ \phi(x_j)=W_\phi x_j
\end{split}
\end{equation}
%------------------------------------------------------------------------

Here $W_g,W_\theta,W_\phi$ is implemented as $1\times1$ convolution in space. Eq.(3) and Eq.(4) represent Gaussian and Embedded Gaussian respectively. $\frac{1}{C(x)}\ f(x_i,\ x_j)$ becomes the \textit{softmax} computation along the dimension $j$. We implement Eq.(4) In the spatial attention module and Eq.(3) in the channel attention module.

The main operations are the same in the two domains' attention module. As shown in Fig.\ref{mab}, given the $F_2\in\mathbb{R}^{C\times H\times W}$, we generate new feature maps $S_1$, $S_2$, $C_1$, and $C_2$, where $\left\{S_1,\ C_1\right\}\in\ \mathbb{R}^{C\times HW}$ and $\left\{S_2,\ C_2\right\}\in\mathbb{R}^{HW\times C}$. Then we perform matrix multiplication and generate the spatial attention map $S_3\in\mathbb{R}^{HW\times HW}$, the channel attention map $C_3\in\mathbb{R}^{C\times C}$ respectively. The process can be formulated as follows:

\begin{equation}
\begin{split}
S_{3}^{ji} = ~\frac{exp\left( {S_{1}^{i} \cdot S_{2}^{j}} \right)}{{\sum\limits_{i = 1}^{HW}{exp}}\left( {S_{1}^{i} \cdot S_{2}^{j}} \right)} 
\end{split}
\end{equation}
\begin{equation}
\begin{split}
C_{3}^{ji} = ~\frac{exp\left( {C_{1}^{i} \cdot C_{2}^{j}} \right)}{{\sum\limits_{i = 1}^{C}{exp}}\left( {C_{1}^{i} \cdot C_{2}^{j}} \right)}
\end{split}
\end{equation}

Where $S_3^{ji}$ and $C_3^{ji}$ denotes the $i$-th spatial/channel’s influence on $j$-th spatial/channel. For the final sum or multiplication, we scale the output by a learnable factor and define the output of spatial/channel as below.

\begin{equation}
\begin{split}
S_{final} = ~\eta{\sum\limits_{i = 1}^{HW}{\left( {S_{3}^{ji} \cdot S_{4}^{i}} \right){\cdot M}_{s}^{j}}}
\end{split}
\end{equation}
\begin{equation}
\begin{split}
C_{final} = ~\mu{\sum\limits_{i = 1}^{C}{\left( {C_{3}^{ji} \cdot C_{4}^{i}} \right) + F_{c}^{j}}}
\end{split}
\end{equation}

$F_3$, the output of our MAB, is generated by integrating the two types of feature maps and the 1×1 convolution. The output is as the same size of $F_2$.

\subsection{Two-stage Training Strategy}
Note that the ability of the image fusion strategy in our framework to reconstruct the fusion image is absolutely crucial for the unified training process at the beginning, accordingly, we develop a two-stage training strategy to make sure that each part of our network could show outstanding performance.

\begin{figure}[t]
      \centering
      \framebox{{\includegraphics[scale=0.24]{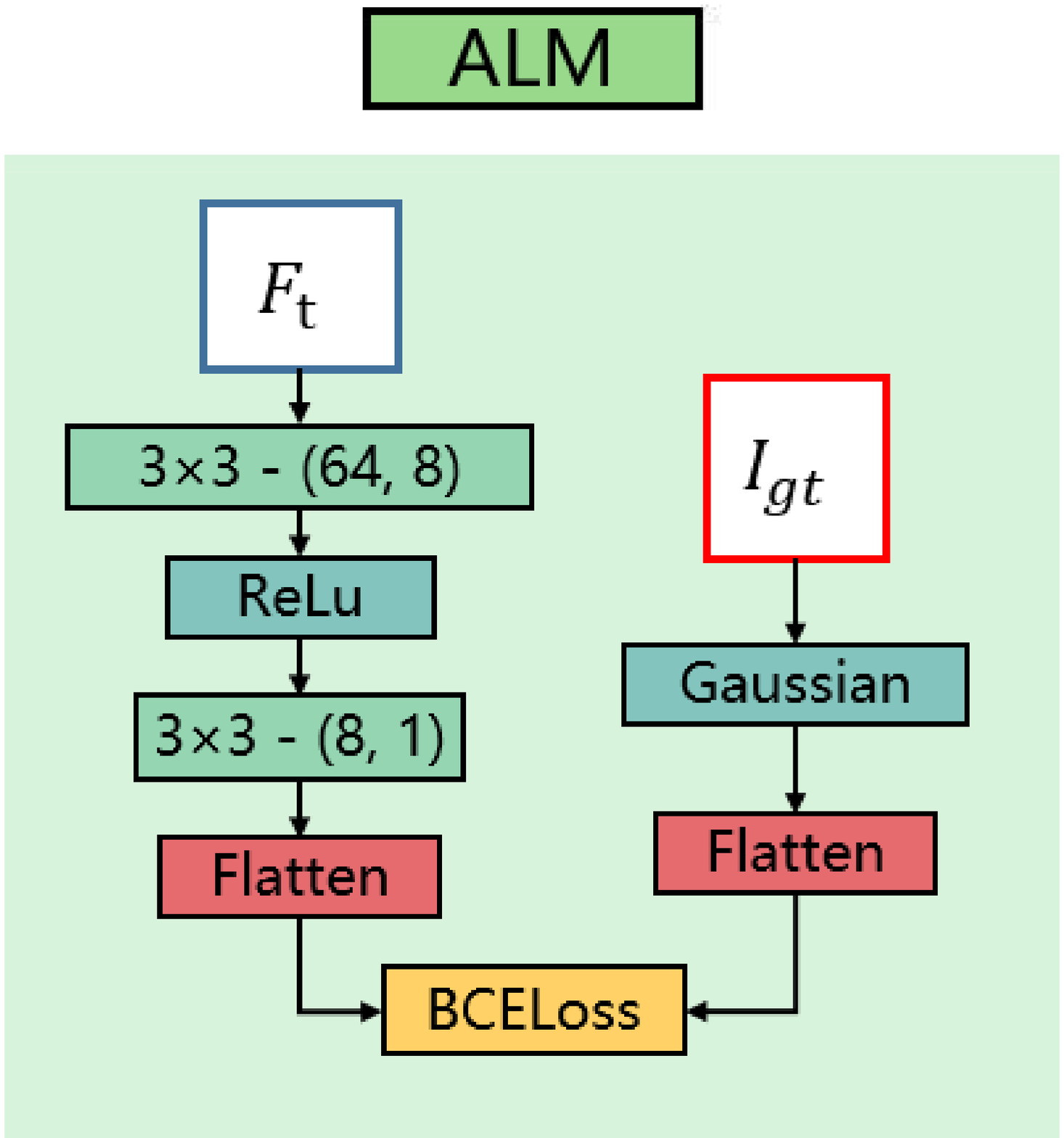}}}
      \caption{The framework of ALM. '3$\times$3 - (64, 8)' represents the convolutional operation with kernel size of 3$\times$3, 64 input channels, 8 output channels.}
      \label{alm}
   \end{figure}

Firstly, we pre-train the CRFN-Nest to extract and reconstruct the multi-modal features. As stated previously, we develop the ALM to guide the fusion model to embody the counting density map features. By using the feature map $F_{t}$ generated by CONV1 in Fig.\ref{wholemodel} and the dot ground truth $I_{g t}$, ALM calculates the classification loss between density map ground truth $k$ and feature map $u$ following Eq.\ref{lossa1} and Eq.\ref{lossa2}. $N$ means the number of images and $c$ means one of the two kinds of modalities (visible or infrared). To generate the density map ground truth $D(x)$, we follow CSRNet by blurring each head annotation using a Gaussian kernel (which is normalized to 1). The geometry-adaptive kernel is defined as Eq.\ref{gs}. Also, $S(x)$ represents the neural network which is shown in the left part of Fig.\ref{alm}. 

\begin{equation}
\begin{split}
\operatorname{Los} s_{A_{t}}=-\frac{1}{N} \sum_{i=1}^{N} k_{i} \log \left(u_{i}\right)+\left(1-k_{i}\right) \log \left(1-u_{i}\right)
\label{lossa1}
\end{split}
\end{equation}

\begin{equation}
\begin{split}
k=D\left(I_{g t}\right), u=S\left(F_{t}\right),\ t \in\{i r, v i\}
\label{lossa2}
\end{split}
\end{equation}

\begin{equation}
\begin{split}
D(x)=\sum_{i=1}^{C} \delta\left(x-x_{i}\right) * G_{\sigma}(x)
\label{gs}
\end{split}
\end{equation}

As the infrared image contains more salient target features than the visible image, the loss function $Loss_{feature}$ is designed to constrain the fused deep features to preserve the salient structures. We set $Loss_{feature}$ following RFN-Nest\cite{RFN}, In Eq.\ref{rfn}, $M$ is the number of the multi-scale deep features, which is set to 4, $w_1$ is a trade-off parameter vector for balancing the loss magnitudes. It assumes four values {1, 10, 100, 1000}. $w_{vi}$and $w_{ir}$ control the relative influence of the visible and infrared features in the fused feature map $\Phi_f^m$. Finally, we set the $Loss_{f}$ defined as Eq.\ref{lossf}.
\begin{equation}
\begin{split}
\text { Loss }_{\text {feature }}=\sum_{j=1}^{M} w_{1}(m)\left\|\Phi_{f}^{m}-\left(w_{v i} \Phi_{v i}^{m}+w_{i r} \Phi_{i r}^{m}\right)\right\|_{F}^{2}
\label{rfn}
\end{split}
\end{equation}

\begin{equation}
\begin{split}
\operatorname{Loss}_{f}=\operatorname{Loss}_{A_{i r}}+\operatorname{Loss}_{A_{v i}}+\text { Loss }_{\text {feature }}
\label{lossf}
\end{split}
\end{equation}

Lastly, we use the pre-trained image fusion model to further train the MFCC to complete the unified training process. During the training phase, the counting loss function $Loss_{counting}$ is standard Mean Squared Error (MSE). The total loss function is defined as Eq.\ref{totalloss}. And the $\lambda$ and $\mu$ are the weights of counting loss and image fusion loss, respectively.
\begin{equation}
\begin{split}
\operatorname{Loss}_{f}=\lambda * \operatorname{Loss}_{\text {counting }}+\mu * \operatorname{Loss}_{f}
\label{totalloss}
\end{split}
\end{equation}

\subsection{The Dense Area Supervising Method}

\begin{figure}[t]
      \centering
      \framebox{\parbox[c][3.3cm][c]{3in}{\includegraphics[scale=0.26]{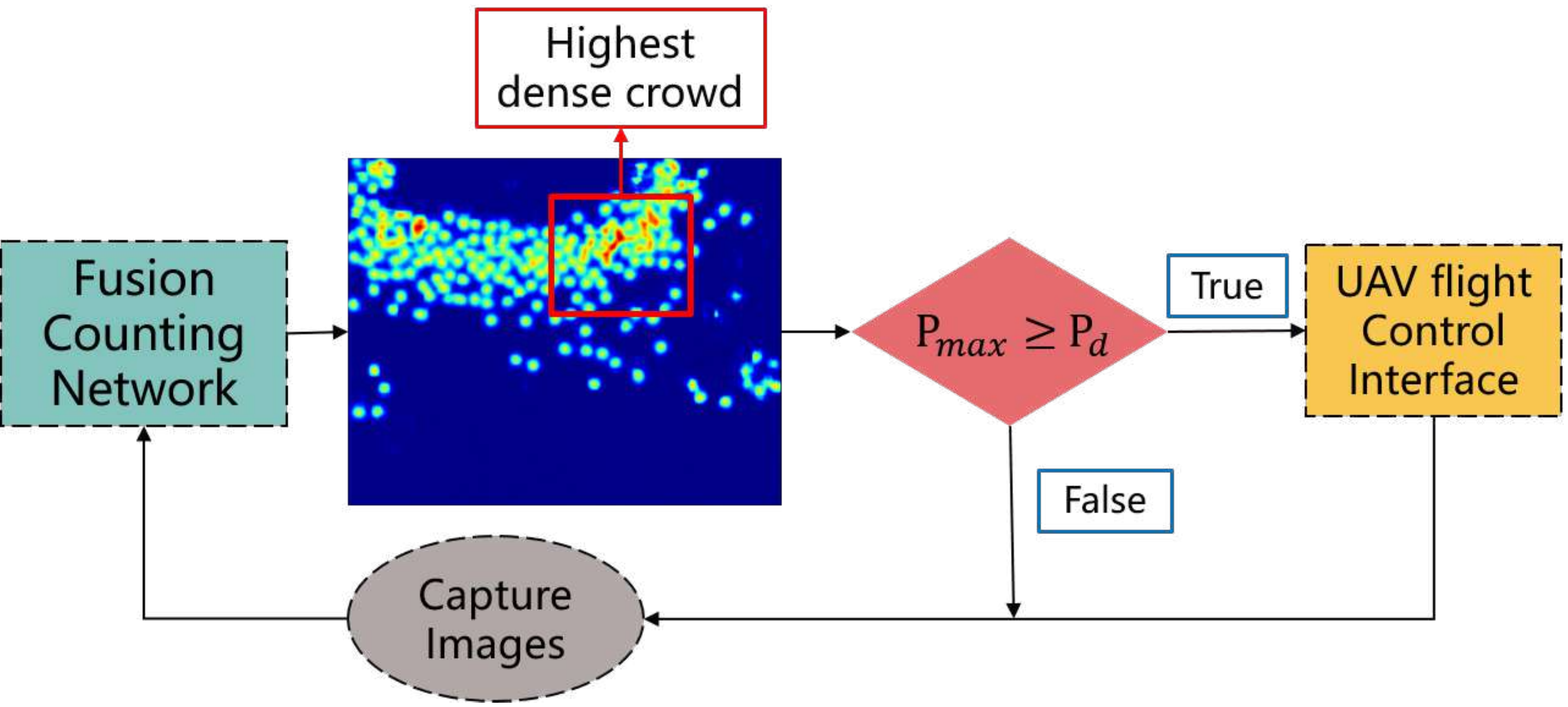}}}
      \caption{The pipeline of dense area supervising method.}
      \label{drone}
   \end{figure}

In addition to design the end-to-end Multi-Task Fusion Counting Network, we also want the prediction map can guide the flight direction of the drone in order to add the ability of real-time supervising the high dense crowds to drones. 
Firstly, we set the dense warning criteria $P_d$ to measure the densest crowd in the prediction image depending on the phenomena. Then we use the window selection algorithm to get several candidate boxes $b_k$ and calculate the total number of people in $b_k$ following Eq.\ref{bk} to get the maximum value $P_{max}$. The function $Count(x)$ means the total number of people included in $b_k$. Secondly, we compare $P_{max}$ and $P_d$ to decide whether the most dense crowd exceeds the dense warning criteria and whether to send the alert message to the UAV flight control interface to supervise the highest density area. The alert message contains the crowd intensity and dense area direction according to the selected $b_k$. Finally, the drone will move directly above the selected $b_k$ by using GPS information of the drone and relative location Information to calculate the transformation matrix and then complete the supervising task, referring \cite{drone flight}. The flow chart is shown in Fig.\ref{drone}. 

\begin{equation}
\begin{split}
\mathrm{P}_{\max }=\operatorname{Max}\left(\operatorname{Count}\left(\mathrm{b}_{k}\right)\right)
\label{bk}
\end{split}
\end{equation}

\begin{figure}[t]
      \centering
      \framebox{\parbox[c][3.2cm][c]{3.2in}{\includegraphics[scale=0.28]{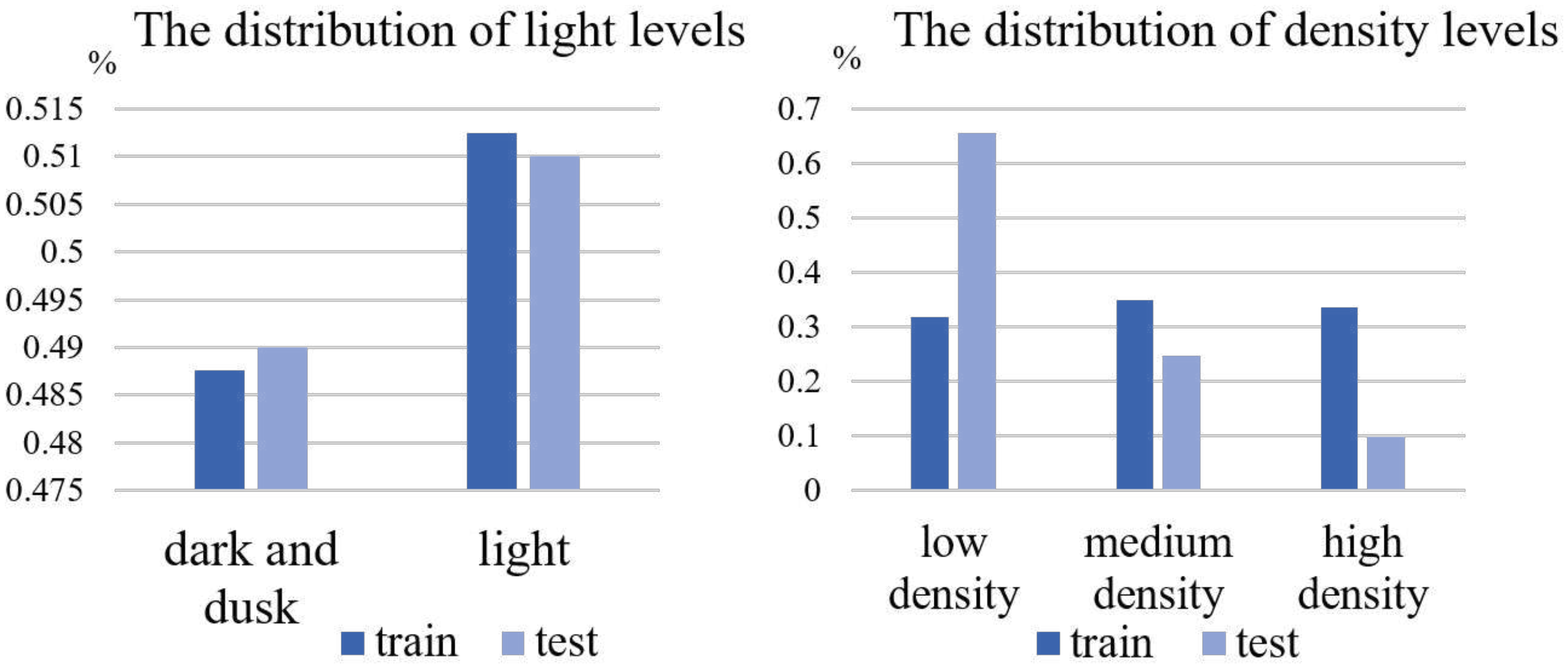}}}
      \caption{The distribution of different light levels and density levels in DroneRGBT data set.}
      \label{distribution}
   \end{figure}

\begin{table*}
\caption{The ablation experiment based on DroneRGBT data set}
\centering
\begin{tabular}{|c|c|c|c|c|c|c|c|c|c|c|c|c|} 
\hline
~ ~ Method & \multicolumn{2}{c|}{Overall} & \multicolumn{2}{c|}{Low density} & \multicolumn{2}{c|}{Medium density} & \multicolumn{2}{c|}{High density} & \multicolumn{2}{c|}{Dark and dust} & \multicolumn{2}{c|}{Light}  \\ 
\hline
Evaluation & MAE  & RMSE                   & MAE  & RMSE                       & MAE   & RMSE                         & MAE   & RMSE                       & MAE  & RMSE                         & MAE   & RMSE                 \\ 
\hline
ALM        & 9.48 & 15.16                 & 5.08 & 7.20                      & 14.17 & 17.63                       & 27.00 & 34.84                     & 8.08 & 13.51                       & 10.83 & 16.59               \\ 
\hline
ECEM+MAB   & 8.28 & 12.76                 & 4.61 & 6.66                      & 14.67 & 18.79                       & 16.72 & 21.79                     & 8.42 & 13.81                       & 8.15  & 11.67               \\ 
\hline
MFCC       & 7.96 & 12.50                 & 4.44 & 6.72                      & 14.09 & 18.68                       & 16.01 & 20.32                     & 7.68 & 13.56                       & 8.30  & 11.51               \\
\hline
\end{tabular}\label{ablation}
\end{table*}

\section{Experiment}
In our experiments, all images are kept the original size 512×640, and the ground truth of density maps is generated under the same size. We set the learning rate at $10^{-5}$ and reduces to 0.995 times every epoch. The batch size of each dataset is 4 on each GPU. We obtain the best result at about 200 epochs. In the MAB, we set the input channel number $C$ as 64, which is the channel number of the output of ECEM. All experiment training and evaluation are performed on NVIDIA GTX 3090 GPU using the PyTorch framework\cite{pytorch}. The backbone of our framework is ResNet-50 and $\lambda$, $\mu$ in Eq.\ref{loss} are set by 10 and 1. We add several Drop-out layers to overcome the over fitting problem.

\subsection{Data set and Evaluation methods} 
In order to assess the performance of the proposed framework, since the multi-modal crowd counting data set is rare, we validate our proposal on DroneRGBT\cite{RGBT} data set. The detailed information of the data set is shown in Table.\ref{dataset}, in which ‘Min’, ’Max’, and ‘Ave’ denote the minimum, maximum, and the average number of people contained in each image in DroneRGBT. We also show the distribution of different illumination levels and density levels in Fig.\ref{distribution}. We evaluate the performance by using Mean Absolute Error (MAE) and Root Mean Square Error (RMSE), which are matrixed as follow:
\begin{equation}
\begin{split}
M A E=\frac{1}{N} \sum_{i=1}^{N}\left|y_{i}-\hat{y}_{i}\right|, R M S E=\sqrt{\frac{1}{N} \sum_{i=1}^{N}\left|y_{i}-\hat{y}_{i}\right|^{2}}
\label{evaluation}
\end{split}
\end{equation}

where $N$ is the number of images in the testing set, $y_i$ is the ground truth of people number and $\hat{y_i}$ refers to the estimated count value for the $i$-th test image.

\begin{table}[h]
\caption{The detailed information of DroneRGBT data set.}
\centering
\begin{tabular}{|c|c|c|c|c|c|c|} 
\hline
Data set   \& Resolution \& Thermal \& View  \& Max \& Min \& Ave    \\ 
\hline
DroneRGBT \& 512×640    \& contain       \& drone \& 403 \& 1   \& 48.8   \\
\hline
\end{tabular}\label{dataset}
\end{table}

\subsection{Compared with baselines}
\begin{table}
\caption{Comparison of our approach with other proposed baseline on DroneRGBT data set}
\centering
\begin{tabular}{|c|c|c|} 
\hline
Method                & \multicolumn{2}{c|}{RGB Input}          \\ 
\hline
~                     & MAE   & RMSE                            \\ 
\hline
ACSCP(CVPR2018)\cite{ACSCP}     & 18.87 & 28.83                           \\ 
\hline
CSRNet(CVPR2018)\cite{csr}    & 13.06 & 19.06                           \\ 
\hline
CANNET(CVPR2019)\cite{CANNET}    & 10.87 & 17.58                           \\ 
\hline
BL(CVPR2019)\cite{BL}        & 10.9  & 16.80                           \\ 
\hline
~                     & \multicolumn{2}{c|}{Thermal Input}      \\ 
\hline
SANET(ECCV 2018)\cite{SANET}      & 12.13 & 17.52                           \\ 
\hline
DA-NET(Access 2018)\cite{DA-NET}   & 9.41  & 14.10                           \\ 
\hline
SCAR(NeuCom 2019)\cite{scar}     & 8.21  & 13.12                           \\ 
\hline
~                     & \multicolumn{2}{c|}{Multi-modal Input}  \\ 
\hline
DroneRGBT(ACCV2020)\cite{RGBT} & 7.27  & 11.45                           \\ 
\hline
MFCC(Ours)            & 7.96  & 12.50                           \\
\hline
\end{tabular}
\label{baseline}
\end{table}

Firstly, compared with the performance of the state-of-art models on RGB mode, our framework has nearly 3.0 percentage improvements on MAE and 4.3 percentage improvements on MSE which proves the effectiveness of our framework. Secondly, we also get improvements compared with the results of the models on thermal infrared mode. Thirdly, the performance of MMCCN proposed along with DroneRGBT data set has a slight advantage over ours because it utilizes CycleGan\cite{Cyclegan} to generate extra fake multi-modal images to improve the accuracy of the prediction density map which consumes more computation compared with ours. Thus, according to the experiments compared with the different baselines, our framework proves its effectiveness and competitiveness.

\subsection{Ablation Experiment}
To analyze the importance of each component of our proposed MFCC, we additionally construct the ablation experiment to prove the effectiveness of each component. In addition to the show the performance of the total test set on objective evaluations, we also compared them under different density levels and light conditions which is consistent with Fig.\ref{distribution}. All components are trained on the training set and tested on the testing set. From Table.\ref{ablation}, our MFCC achieves better results regardless of the density and light condition.

\begin{figure}[t]
      \centering
      \framebox{{\includegraphics[scale=0.5]{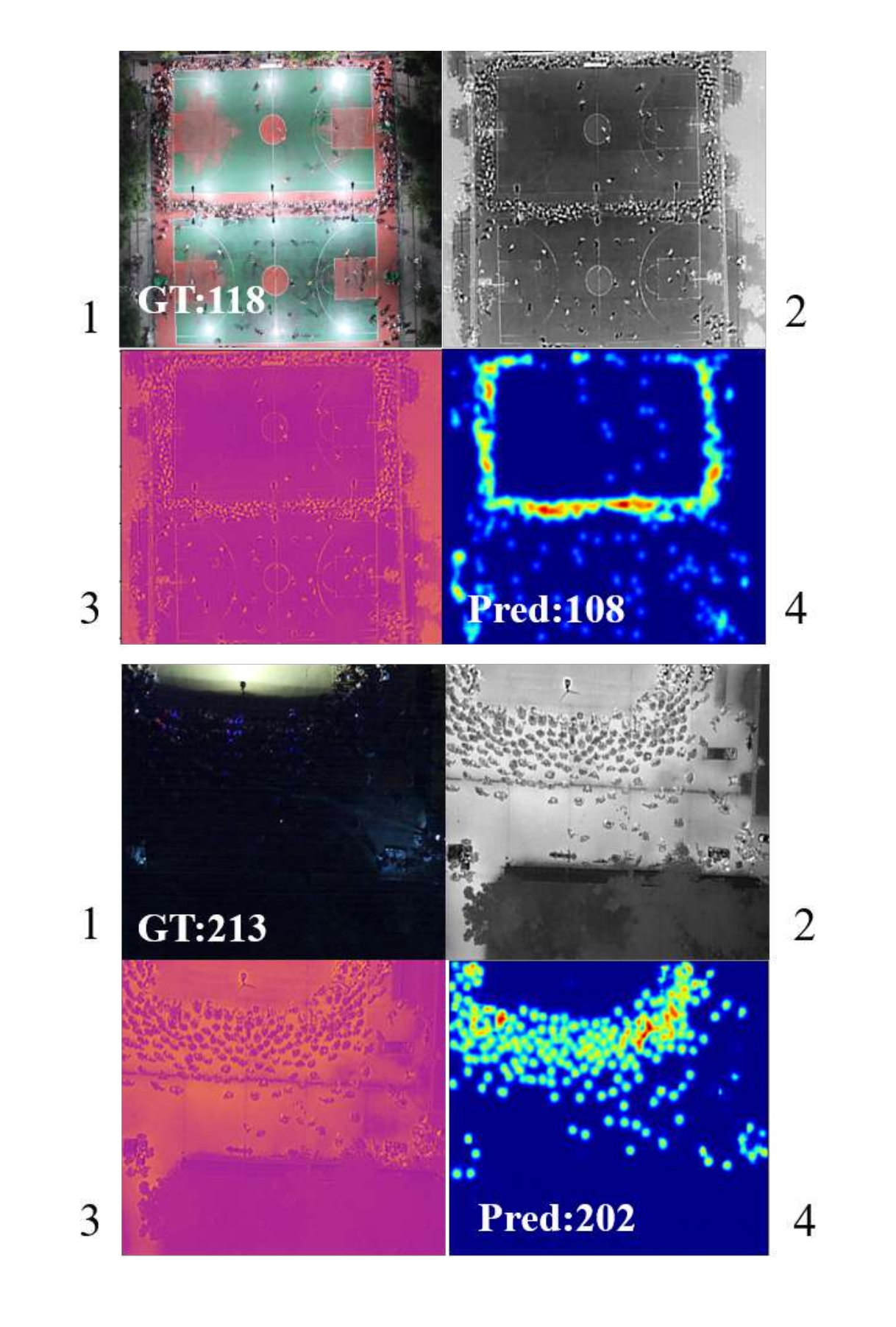}}}
      \caption{The visualization of estimating the result of DroneRGBT  data set. Pictures with number 1 represent visible image, pictures with number 2 represent thermal infrared image, pictures with number 3 represent fusion image, pictures with number 4 represent prediction map. 'GT' represents the ground truth population and 'Pred' represents the estimation population.}
      \label{test}
   \end{figure}

To intuitively show the performance of our method, Fig.\ref{test} illustrates the 2 groups of visualization results on the DroneRGBT data set. We can clearly find that the prediction counting numbers of our framework approaches to the label counting numbers.

\section{Conclusion}
In this paper, we propose a Unified Multi-Task Learning Framework of Real-Time Drone Supervision for Crowd Counting (MFCC) to estimate the density map and crowd counting result, and then utilize the prediction map for drones flight to supervise the dense crowd. We propose this framework for the first time to prove the effectiveness of our method through different experiments settings. We will enhance our framework to achieve better results in the feature and design more efficient models.


%%%%%%%%%%%%%%%%%%%%%%%%%%%%%%%%%%%%%%%%%%%%%%%%%%%%%%%%%%%%%%%%%%%%%%%%%%%%%%%%

%%%%%%%%%%%%%%%%%%%%%%%%%%%%%%%%%%%%%%%%%%%%%%%%%%%%%%%%%%%%%%%%%%%%%%%%%%%%%%%%

%%%%%%%%%%%%%%%%%%%%%%%%%%%%%%%%%%%%%%%%%%%%%%%%%%%%%%%%%%%%%%%%%%%%%%%%%%%%%%%%

%%%%%%%%%%%%%%%%%%%%%%%%%%%%%%%%%%%%%%%%%%%%%%%%%%%%%%%%%%%%%%%%%%%%%%%%%%%%%%%%

\addtolength{\textheight}{-12cm}


\begin{thebibliography}{99}
\bibitem{mcnn} Zhang Y, Zhou D, Chen S, et al. Single-image crowd counting via multi-column convolutional neural network[C]//Proceedings of the IEEE conference on computer vision and pattern recognition. 2016: 589-597
\bibitem{VLAD}Sheng B, Shen C, Lin G, et al. Crowd counting via weighted VLAD on a dense attribute feature map[J]. IEEE Transactions on Circuits and Systems for Video Technology, 2016, 28(8): 1788-1797
\bibitem{scar}Gao, J., Wang, Q., \& Yuan, Y. (2019). SCAR: Spatial-/channel-wise attention regression networks for crowd counting. Neurocomputing, 363, 1–8
\bibitem{csr}Li Y, Zhang X, Chen D. Csrnet: Dilated convolutional neural networks for understanding the highly congested scenes[C]//Proceedings of the IEEE conference on computer vision and pattern recognition. 2018: 1091-1100
\bibitem{DM}Wang, B., Liu, H., Samaras, D., \& Hoai, M. (2020). Distribution Matching for Crowd Counting. NeurIPS, 1–13
\bibitem{noisy}Wan J, Chan A. Modeling noisy annotations for crowd counting[J]. Advances in Neural Information Processing Systems, 2020, 33.
\bibitem{drone-scnet}Elharrouss O, Almaadeed N, Abualsaud K, et al. Drone-SCNet: Scaled Cascade Network for Crowd Counting on Drone Images[J]. IEEE Transactions on Aerospace and Electronic Systems, 2021
\bibitem{geometric}Liu, W., Lis, K., Salzmann, M., \& Fua, P. (2019). Geometric and Physical Constraints for Drone-Based Head Plane Crowd Density Estimation. IEEE International Conference on Intelligent Robots and Systems, 244–249
\bibitem{scale-adaptive}Küchhold M, Simon M, Eiselein V, et al. Scale-adaptive real-time crowd detection and counting for drone images[C]//2018 25th IEEE International Conference on Image Processing (ICIP). IEEE, 2018: 943-947
\bibitem{soft-csrnet}Bakour, I., Bouchali, H. N., Allali, S., \& Lacheheb, H. (2021). Soft-CSRNet: Real-time Dilated Convolutional Neural Networks for Crowd Counting with Drones. 2020 2nd International Workshop on Human-Centric Smart Environments for Health and Well-Being, IHSH 2020, 28–33
\bibitem{drone1}Sirmacek B, Reinartz P. Automatic crowd analysis from airborne images[C]//Proceedings of 5th International Conference on Recent Advances in Space Technologies-RAST2011. IEEE, 2011: 116-120
\bibitem{drone2}Gonzalez-Trejo J, Mercado-Ravell D. Dense Crowds Detection and Surveillance with Drones using Density Maps[C]//2020 International Conference on Unmanned Aircraft Systems (ICUAS). IEEE, 2020: 1460-1467
\bibitem{drone3}Tzelepi M, Tefas A. Human crowd detection for drone flight safety using convolutional neural networks[C]//2017 25th European Signal Processing Conference (EUSIPCO). IEEE, 2017: 743-747
\bibitem{RFN}Li H, Wu X J, Kittler J. RFN-Nest: An end-to-end residual fusion network for infrared and visible images[J]. Information Fusion, 2021, 73: 72-86
\bibitem{nestfuse}Li H, Wu X J, Durrani T. NestFuse: An infrared and visible image fusion architecture based on nest connection and spatial/channel attention models[J]. IEEE Transactions on Instrumentation and Measurement, 2020, 69(12): 9645-9656
\bibitem{Did}Zhao Z, Xu S, Zhang C, et al. DIDFuse: Deep image decomposition for infrared and visible image fusion[J]. arXiv preprint arXiv:2003.09210, 2020.

\bibitem{Res50}He K, Zhang X, Ren S, et al. Deep residual learning for image recognition[C]//Proceedings of the IEEE conference on computer vision and pattern recognition. 2016: 770-778
\bibitem{Attention-guided}Cao, J., Chen, Q., Guo, J., \& Shi, R. (2020). Attention-guided context feature pyramid network for object detection. ArXiv, 1–12
\bibitem{DenseNet}Huang G, Liu Z, Van Der Maaten L, et al. Densely connected convolutional networks[C]//Proceedings of the IEEE conference on computer vision and pattern recognition. 2017: 4700-4708
\bibitem{Non-local}Wang X, Girshick R, Gupta A, et al. Non-local neural networks[C]//Proceedings of the IEEE conference on computer vision and pattern recognition. 2018: 7794-7803
\bibitem{local-aware}Zhou J T, Zhang L, Jiawei D, et al. Locality-Aware Crowd Counting[J]. IEEE Transactions on Pattern Analysis and Machine Intelligence, 2021.
\bibitem{drone flight}Skoda J, Barták R. Camera-based localization and stabilization of a flying drone[C]//The Twenty-Eighth International Flairs Conference. 2015
\bibitem{RGBT}Peng, T., Li, Q., \& Zhu, P. (2021). RGB-T Crowd Counting from Drone: A Benchmark and MMCCN Network. Lecture Notes in Computer Science (Including Subseries Lecture Notes in Artificial Intelligence and Lecture Notes in Bioinformatics), 12627 LNCS, 497–513 
\bibitem{pytorch}Paskze A, Chintala S. Tensors and Dynamic neural networks in Python with strong GPU acceleration[J]. 2017
\bibitem{ACSCP}Shen Z, Xu Y, Ni B, et al. Crowd counting via adversarial cross-scale consistency pursuit[C]//Proceedings of the IEEE conference on computer vision and pattern recognition. 2018: 5245-5254
\bibitem{CANNET}Liu W, Salzmann M, Fua P. Context-aware crowd counting[C]//Proceedings of the IEEE/CVF Conference on Computer Vision and Pattern Recognition. 2019: 5099-5108
\bibitem{SANET}Z. Shen, Y. Xu, B. Ni, M. Wang, J. Hu and X. Yang, "Crowd Counting via Adversarial Cross-Scale Consistency Pursuit," 2018 IEEE/CVF Conference on Computer Vision and Pattern Recognition, 2018, pp. 5245-5254
\bibitem{DA-NET}Zou Z, Su X, Qu X, et al. Da-net: Learning the fine-grained density distribution with deformation aggregation network[J]. IEEE Access, 2018, 6: 60745-60756
\bibitem{BL}Ma Z, Wei X, Hong X, et al. Bayesian loss for crowd count estimation with point supervision[C]//Proceedings of the IEEE/CVF International Conference on Computer Vision. 2019: 6142-6151
\bibitem{Cyclegan}Zhu J Y, Park T, Isola P, et al. Unpaired image-to-image translation using cycle-consistent adversarial networks[C]//Proceedings of the IEEE international conference on computer vision. 2017: 2223-2232
\end{thebibliography}
\end{document}